\documentclass[lettersize,journal,twoside]{IEEEtran}
\usepackage{amsmath,amsfonts}
\usepackage{algorithmic}
\usepackage{algorithm}
\usepackage{array}
\usepackage[caption=false,font=normalsize,labelfont=sf,textfont=sf]{subfig}
\usepackage{textcomp}
\usepackage{stfloats}
\usepackage{url}
\usepackage{verbatim}
\usepackage{graphicx}
\usepackage{cite}
\usepackage{amsmath,amssymb,amsfonts}
\usepackage{multirow}
\usepackage{algorithmic}
\usepackage{graphicx}
\usepackage{float}
\usepackage{subfig}
\usepackage{makecell}
\usepackage{booktabs}
\usepackage{multirow}
\usepackage{textcomp}
\usepackage{color}
\usepackage{balance}
\usepackage[normalem]{ulem}
\useunder{\uline}{\ul}{}
\usepackage{ragged2e}

\begin{document}

\title{TrustGNN: Graph Neural Network based Trust Evaluation via Learnable Propagative and Composable Nature}

\author{Cuiying Huo,
        ~Di Jin,
        ~Chundong Liang,
        ~Dongxiao He,
        ~Tie Qiu
        and~Lingfei Wu

\thanks{Cuiying Huo, Di Jin,  Chundong Liang, Dongxiao He and Tie Qiu are with the College of Intelligence and Computing, Tianjin University, Tianjin 300350, China.  E-mail: huocuiying@tju.edu.cn; jindi@tju.edu.cn; liangchundong@tju.edu.cn; hedongxiao@tju.edu.cn; qiutie@ieee.org.}

\thanks{Lingfei Wu is with the JD.COM Silicon Valley Research Center, 675 E Middlefield Rd, Mountain View, CA 94043 USA. E-mail:lwu@email.wm.edu.}

}

\maketitle

\begin{abstract}
Trust evaluation is critical for many applications such as cyber security, social communication and recommender systems. Users and trust relationships among them can be seen as a graph. Graph neural networks (GNNs) show their powerful ability for analyzing graph-structural data. Very recently, existing work attempted to introduce the attributes and asymmetry of edges into GNNs for trust evaluation, while failed to capture some essential properties (e.g., the propagative and composable nature) of trust graphs. In this work, we propose a new GNN based trust evaluation method named TrustGNN, which integrates smartly the propagative and composable nature of trust graphs into a GNN framework for better trust evaluation. Specifically, TrustGNN designs specific propagative patterns for different propagative processes of trust, and distinguishes the contribution of different propagative processes to create new trust. Thus, TrustGNN can learn comprehensive node embeddings and predict trust relationships based on these embeddings. Experiments on some widely-used real-world datasets indicate that TrustGNN significantly outperforms the state-of-the-art methods. We further perform analytical experiments to demonstrate the effectiveness of the key designs in TrustGNN.
\end{abstract}


\section{Introduction}\label{sec:introduction}
With the evolution of communication technology and the widespread popularity of the Internet, the number of social network users is growing rapidly. Online social networks such as Facebook, LinkedIn and Twitter have become an integral part of their user’s daily life. However, due to the inherent openness of online social networks, anyone can join these networks, which inevitably provides an opportunity for malicious users to spread incorrect or illegal information~\cite{illegal1,illegal2}. Therefore, evaluating social trust, which plays a crucial role in the functionality and operation of social networks, has become an important topic in online social network analysis.

\begin{figure}[t]  
  \centering       
  \includegraphics[width=0.9\linewidth]{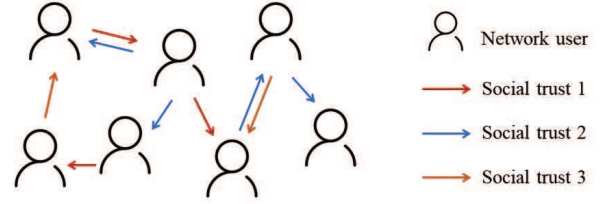} 
  \caption{The social network schema based on social trust relationships. The arrow points from the trustor to the trustee. Different colored arrows represent different types of social trust relationships.}
  \label{pic:dataset}    
\end{figure}

Trust is the extent by which one user (trustor) expects that another user (trustee) performs a given action~\cite{MATRI}. Trust evaluation is to evaluate the pairwise trust relationship between two users who are directly or indirectly connected within online social networks. As shown in Fig.~\ref{pic:dataset}, trust-based online social networks usually contain multiple types of social trust relationships. The trust relationship is usually asymmetric, that is, there may be two trust relationships in opposite directions between two users. There are many different attempts proposed to evaluate the social trust in online social networks~\cite{sv,sv1,sv2}. E.g., the subjective logic-based methods~\cite{TNA-SL,3VSL,OpinionWalk}, which follow the assumptions of cognitive recognition and introduce the uncertainty inference process for the subjective nature of trust. The probability statistics-based methods~\cite{Liu1,Itrust,Liu2}, rely on statistical distributions to represent and model social trust relationship in a computational way. The machine learning-based methods~\cite{MATRI,NeuralWalk}, use some machine learning techniques such as matrix factorization to model the trust evaluation task as a learnable problem. However, these existing approaches usually have high computational complexity, or have poor performance because they do not consider the user's attribute information.

In recent years, with the great success of deep learning, Graph Neural Networks (GNNs)~\cite{gnnsurvey1, gnnsurvey2,book}, as powerful tools for processing graph data, have shown superior performance on various network analysis tasks, such as node classification~\cite{DropEdge,AM-GCN}, link prediction~\cite{link_prediction1,link_prediction12} and recommendation~\cite{Recommendation1,BasConv}. The essence of GNNs is the process of information propagation and aggregation guided by graph structure. These GNN-based methods obtain meaningful representation of nodes/edges in a network by integrating the neighborhood information of nodes/edges. On the other hand, an online social network based on social trust can be thought of as a trust graph where the nodes are social users and the edges are the trust relations between them. The edge in the graph can represent the trust relationship between two nodes. Therefore, it is significant to utilize the powerful representation learning capabilities of GNN for trust evaluation tasks.

 To our best knowledge, there has been only one attempt to apply GNNs to trust evaluation in online social networks, i.e., Guardian~\cite{Guardian}, presented very recently. Guardian divides the neighbor nodes into the set of in-degree neighbor nodes and the set of out-degree neighbor nodes, and uses a graph convolutional network~\cite{gcn} (GCN) layer for information aggregation respectively. It uses an mean aggregator to aggregate the information of different first-order neighbors, and aggregates high-order neighbor information by stacking multiple GCN layers. But the design of Guardian does not take full account of the nature of social trust in social networks, e.g., the propagative and composable nature, which are however essentially important for trust evaluation. To sum up, in order to better apply GNNs to trust evaluation, there are three main challenges:
\begin{enumerate}
    \item The propagative nature indicates that trust can be passed between users along a chain, so we can obtain the trust value between any two users who are indirectly connected through the chain. Guardian fails to explicitly consider the role of the social trust chain, which will result in redundancy or lack of effective information. In this context, the first challenge is how we can explicitly integrate the trust chain into GNN frameworks in order to better model the propagative nature.
    \item Due to the asymmetry of the trust relationship, the directionality of trust propagation also needs to be considered. Simply grouping neighbor nodes based on in-degree and out-degree is not suitable for trust chains. Therefore, the second challenge is how to define a directional propagation pattern for trust chains.
    \item The composable nature indicates us that when there are multiple trust chains between two users, the trust value needs to be aggregated by considering the interaction between these chains. The mean aggregator of Guardian cannot distinguish the information from differnet chains. Thus, the third challenge is how to assign different weights according to the importance coefficients of neighbor nodes to consider the composable nature more comprehensively.
\end{enumerate}

Facing the above challenges, we propose a GNN-based trust evaluation method that comprehensively utilizes the nature of social trust, namely TrustGNN. Since there are usually multiple types of trust relationships in online social networks, we first define multiple specific trust chains according to the trust types. The trust chain is composed of multiple users and trust relationships. Inspired by knowledge graph embedding methods, we define the propagative pattern of information on chains by considering the interaction between users and trust relationships in a chain, as well as the directionality of social trust. By doing so, the propagative nature and the asymmetry of social trust can be obtained. Moreover, in online social networks, a target node usually needs to aggregate information from multiple trust chains, i.e., the composable nature of social trust. Considering that different types of trust chains have different impact on the target node, we adopt a learnable attention~\cite{gat, attention2} to learn their contributions for better capturing the composable nature of online social networks.

To summarize, the main contributions of this paper are as follows. First, we discuss and clarify what is essentially necessary for GNN to deal with trust evaluation. And then, we introduce that GNN, as an effective tool for processing graph-structured data, can be used for trust evaluation in a more natural way. Second, we propose a GNN-based online social network trust evaluation method (TrustGNN), which more comprehensively and essentially integrates the propagative, composable, and asymmetric nature of social trust into the same GNN framework. Third, extensive experiments and analyses on two online social networks demonstrate the superiority of the proposed method over the state-of-the-art methods.

The rest of the paper is organized as follows. Section~\ref{sec:PDA} gives the problem definitions and analysis. Section~\ref{sec:method} proposes the new approach TrustGNN. We conduct extensive experiments in Section~\ref{sec:Experiments}. Finally, we discuss related work in Section~\ref{sec:relatedWork} and conclude in Section~\ref{sec:conclusion}.

\section{Problem Definition and Analysis}\label{sec:PDA}
\textbf{Trust Graph.} In this paper, we define trust relationships in the online social network as a directed trust graph $\mathcal{G}=\left( \mathcal{V},\mathcal{E},\mathcal{R},\phi  \right)$, where node set $\mathcal{V}$ represents users and edge set $\mathcal{E}$ represents trust relationships among users. The trust type set $\mathcal{R}$ enumerates all trust relationship types in the graph $\mathcal{G}$. The mapping function $\phi :\mathcal{E}\to \mathcal{R}$ maps observed edges to trust relationship types, so each edge strictly corresponds to a specific trust relationship. Moreover, the trustworthiness is different in different application domain. For example, trustworthiness can be simply classified into two types, i.e., $\mathcal{R}=\{\text{Trust, Distrust}\}$. But in some more complicated situations such as in Advogato\footnote{http://trustlet.org/datasets/advogato/} and Pretty-Good-Privacy\footnote{http://networkrepository.com/arenas} (PGP), the trust relations have four types, i.e., $\mathcal{R}=\{\text{Observer, Apprentice, Journeyer, Master}\}$.

\textbf{Trust Evaluation.} The trust evaluation task is to predict the unobserved trust relationships in a trust graph $\mathcal{G}$. Specifically, given a trust graph $\mathcal{G}=\left( \mathcal{V},\mathcal{E},\mathcal{R},\phi  \right)$, the goal is to train a model $f\left( \cdot  \right)$. For nodes $u\in \mathcal{V}$, $v\in \mathcal{V}$, and edge $\left\langle u,v \right\rangle \notin \mathcal{E}$, the model $f\left( \cdot  \right)$ can predict the trust relationship for the two nodes, i.e., $f\left( \left\langle u,v \right\rangle  \right)\to \mathcal{R}$. Note that trust relationships are directed, and the trust relationship from node $u$ to node $v$ does not equal to the relationship from node $v$ to node $u$. Therefore, $f\left( \left\langle u,v \right\rangle  \right)\ne f\left( \left\langle v,u \right\rangle  \right)$.

\begin{figure}[t]  
  \centering       
    \subfloat[The propagative nature] 
    {
        \begin{minipage}[t]{0.22\textwidth}
            \centering          
            \includegraphics[width=0.98\textwidth]{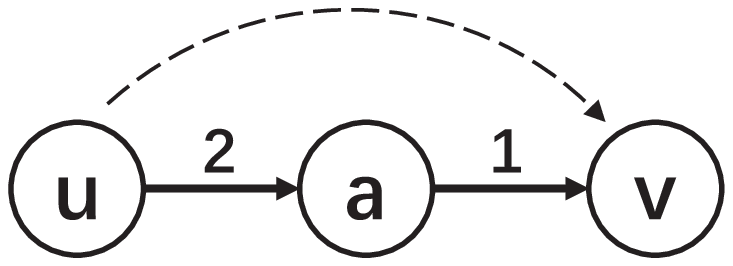}  
        \end{minipage}%
    }
    \subfloat[The composable nature] 
    {
        \begin{minipage}[t]{0.22\textwidth}
            \centering      
            \includegraphics[width=0.98\textwidth]{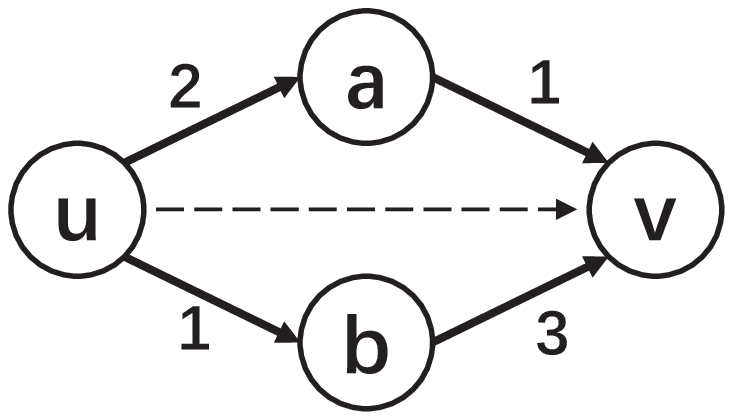}   
        \end{minipage}
    }%
  \caption{The illustrations of properties in trust graphs focused in this work. (a) The propagative nature. Trust is passed from node $u$ to node $v$ along a chain, providing the evidence to create trust relationships between nodes $u$ and $v$. (b) The composable nature. Trust is passed from node $u$ to node $v$ along two chains with different types. The two trust chains that need to be combined to create a new trust relationship between nodes $u$ and $v$ .}\label{fig:properties}  
\end{figure}

\textbf{Properties in Trust Graph.} Before discussing the proposed method, we first show some insights about trust graphs and analyze some properties which are used in our method. For any two nodes $u$ and $v$, there will be a kind of trust relationship from node $u$ to node $v$ if $\left\langle u,v \right\rangle \in \mathcal{E}$. We define $u$ as the \emph{trustor} and $v$ the \emph{trustee}. The two common-used properties in trust evaluation are the propagative nature and composable nature of social trust~\cite{Guardian}. Specifically, \emph{the propagative nature} of social trust means that trust can be passed among nodes, creating trust chains that connects two nodes who are indirectly connected in the graph. For example, in Fig.~\ref{fig:properties}(a), node $u$ trusts node $a$ with a trust value of $2$ and node $a$ trusts node $v$ with a trust value of $1$. There is a \emph{trust chain} between node $u$ and node $v$, which can be taken as evidence to create trust relationship from node $u$ to node $v$. \emph{The composable nature} of social trust means that there may be several trust chains between two nodes, and these chains interact to provide evidence for us to create a new trust relationship between these two nodes (Fig.~\ref{fig:properties}(b)).

In this paper, we define two nodes connected by a trust chain as neighbor nodes. In particular, the head node $u$ is \emph{the trust chain based trustor neighbor} of the tail node $v$, while the tail node $v$ is \emph{the trust chain based trustee neighbor} of the head node $u$.

\begin{table}[t]
  \caption{Notations and Explanations}\label{tab:notation}
  \resizebox{\linewidth}{!}{
  \begin{tabular}{l|c}
    \toprule[1.5pt]
    Notations&Explanations\\
    \midrule[0.6pt]
    ${{x}_{node}}$ & The attributes of a node.\\
    ${{x}_{edge}}$ & The attributes of an edge.\\
    $p$ & A trust chain.\\
    ${{P}_{j}}$ &  The $j$-th type of trust chains.\\
    ${{h}_{v}}$, $H$ & The latent representation of node $v$ / all nodes.\\
    ${{r}_{i}}$ & The latent representation of the $i$-th edge type.\\
    ${{z}_{v}}$, $Z$ & The final node embedding vector of node $v$ / all nodes.\\
    $q$ & The attention vector.\\
    ${{w}_{{{P}_{j}}}}$ & The importance of chain type ${{P}_{j}}$.\\
    ${{\alpha }_{{{P}_{j}}}}$ & The normalized weight of chain type ${{P}_{j}}$.\\
    $W$, $b$ & The parameters of neural networks.\\
    $Y$ & The predicted trust relationships.\\
    $\tilde{Y}$ & The ground truth.\\
    \bottomrule[1.5pt]
  \end{tabular}
  }
\end{table}

\textbf{Graph Neural Networks (GNNs).} GNNs~\cite{gnnsurvey1, gnnsurvey2} are powerful tools for analyzing graph-structural data. A GNN can be interpreted as smoothing local information through propagation and aggregation operations in a graph, and finally learning node representations that can be applied to various downstream tasks. After $k$ iterations of propagation and aggregation, a node’s representation can capture topology and attribute information within its $k$-hop neighborhood in the graph. Formally, in the $k$-th layer, the representation $h_{u}^{(k)}$ of node $u$ is
\begin{equation}\label{eq:gnn-prop}
    a_{v}^{(k)}=\text{Pro}{{\text{p}}^{(k)}}\left( h_{v}^{(k-1)} \right),\quad v\in \mathcal{N}\left( u \right)
\end{equation}
\begin{equation}\label{eq:gnn-agg}
    h_{u}^{(k)}=\text{Ag}{{\text{g}}^{(k)}}\left( h_{u}^{(k-1)},\{a_{v}^{(k)}:v\in \mathcal{N}\left( u \right)\} \right)
\end{equation}
where $\mathcal{N}\left( u \right)$ is the set of the direct neighbors of node $u$, $\text{Pro}{{\text{p}}^{(k)}}\left( \cdot  \right)$ and $\text{Ag}{{\text{g}}^{(k)}}\left( \cdot  \right)$ are two functions implemented by neural networks in the $k$-th GNN layer. $\text{Pro}{{\text{p}}^{(k)}}\left( \cdot  \right)$ corresponds to the node representation transformation in the propagation process and $\text{Ag}{{\text{g}}^{(k)}}\left( \cdot  \right)$ aggregates the transformed neighbor node representations and its own representation.

In this work, we use GNN framework to learn node embeddings and evaluate trust relationships between two nodes based on their embeddings. The key novelty here is that we design a new GNN to simultaneously model propagative and composable nature of trust graphs. Intuitively, different trust chains should have different effects for creating trust. Therefore, there should be different propagation rules on different trust chains in GNNs. In this way, the propagative nature in trust graphs can be well learned. Moreover, when GNNs aggregate neighbor information from different trust chains, it should distinguish the contribution of information from different chains and generate node representations based on their contributions. This is because a trust chain composed of relationships with high value is more helpful to create a new trust relationship. Thus, GNNs can capture the composable nature in trust graphs and learn comprehensive node embeddings for trust evaluation.

In the following we will show the proposed model based on the ideas discussed above. All notations that we used in the paper and their explanations are provided in Table~\ref{tab:notation}.

\section{Method}\label{sec:method}
\subsection{Overview}
\begin{figure*}[t]  
  \centering       
  \includegraphics[width=\linewidth]{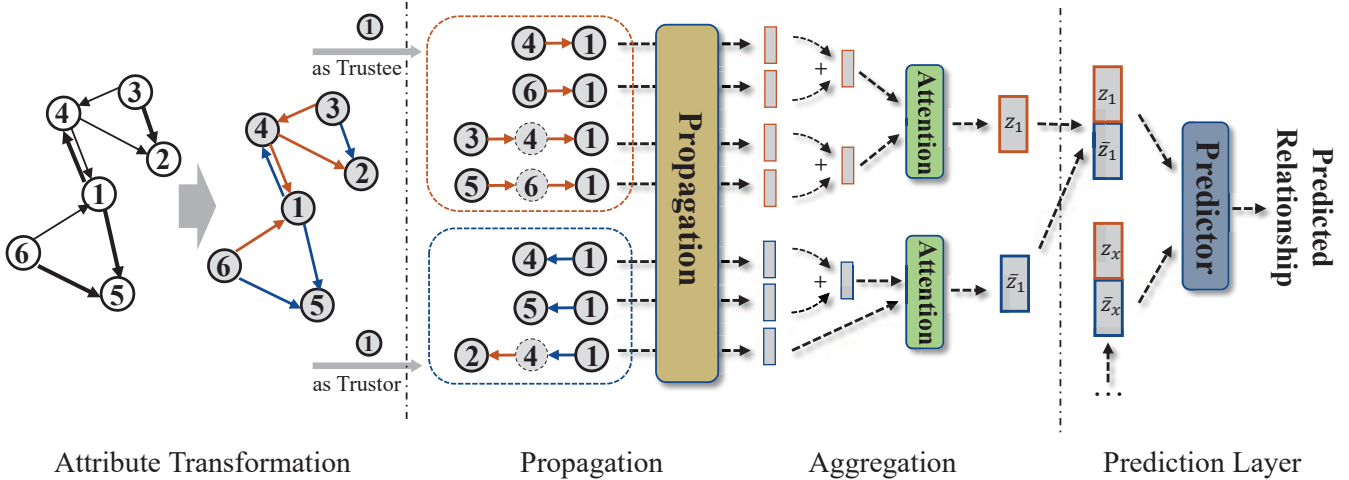} 
  \caption{The flow chart of TrustGNN (take maximum length of the trust chain as 2 as an example). TrustGNN first transforms attributes of nodes and edges into a common vector space, and then propagate and aggregate transformed attributes based on trust chains to generate node embeddings. Due to the asymmetry of trust, TrustGNN generates two kinds of embeddings from the perspective of trustee and trustor, and concatenates them as a final embedding. TrustGNN finally predicts the trust relationships based on embeddings of trustor-trustor node pairs.}\label{fig:overall}  
\end{figure*}

In this section, we present our graph neural network (GNN) based trust evaluation method (TrustGNN). Since the propagative and composable nature is the basis for creating new trust, we integrate these two properties into the GNN framework, and capture them in a learnable manner. The overview of TrustGNN is illustrated in Fig.~\ref{fig:overall}, which consists of four components. i) TrustGNN first initializes the attributes of nodes and edges and projects them into a latent space, i.e., \emph{attribute transformation}. ii) The information is propagated along trust chains. Here TrustGNN extends knowledge graph embedding methods to GNN to better model the propagative pattern of trust, and further considers the asymmetry of trust, i.e., \emph{trust chain based propagation}. iii) TrustGNN identifies the information propagated from different chains and aggregates the information in a discriminable way, and finally generates node embeddings, i.e., \emph{trust chain based aggregation}. iv) TrustGNN uses a multilayer perceptron layer to predict trust relationships based on node pair embeddings, i.e., \emph{the predictor layer}.

\subsection{Attribute Transformation}
In a complete trust graph $\mathcal{G}=\left( \mathcal{V},\mathcal{E},\mathcal{R},\phi  \right)$, there are homogeneous nodes and heterogeneous edges, and nodes and edges both have their attributes. These nodes and edges may have different attribute vector dimensions. Even though they happen to have same dimension, they may still be in different vector spaces. Therefore, we need to transform these attributes into a common vector space to feed to the GNN framework.

TrustGNN uses a linear transformation to transform all node attributes. For each node $v\in \mathcal{V}$, there is
\begin{equation}\label{eq:attr-trans-node}
    {{h}_{v}}={{W}_{node}}\cdot {{x}_{v}}
\end{equation}
where ${{x}_{v}}$ is the attribute vector of node $v$, ${{h}_{v}}$ the transformed representation of node $v$ and ${{W}_{node}}$ a learnable parameter matrix. TrustGNN has different linear transformations corresponding to different types of edges. For the $i$-th type of edge, there is
\begin{equation}\label{eq:attr-trans-edge}
    {{r}_{i}}={{W}_{edge-i}}\cdot {{x}_{edge-i}}
\end{equation}
where ${{x}_{edge-i}}$ is the attribute vector of the $i$-th edge type, ${{W}_{edge-i}}$ a learnable parameter matrix specific to the $i$-th edge type, and ${{r}_{i}}$ the transformed representation of the $i$-th edge type.

Note that not all datasets have completed attributes for nodes and edges. For the datasets missing node or edge attributes, we initialize these missing attributes as random vectors and treat these vectors as a part of learnable parameters in neural networks.

\subsection{Trust Chain Based Propagation}\label{sec:method-prop}
In trust graphs, trust may be passed from one node to another along a trust chain so that two nodes at head and tail of a trust chain can create a new trust relationship, in which the head node is a trustor and the tail node is a trustee. The message passing in GNNs is a similar propagative pattern with trust propagation in a trust graph. Therefore, we can integrate trust chains into GNNs and make GNNs propagate information along trust chains. However, the length of a trust chain can be infinite and there can be infinite types of trust chains in a trust graph. In TrustGNN, we limit the maximum length of the trust chain to the preset hyper-parameter $K$, which is reasonable because trust will decrease in the propagation process and a too long trust chain is not beneficial to creating new trust.

Consider a trustee node $v$ and a trust chain $p$ of length $k$ ($ k\leq K $) $:u\xrightarrow{{{r}_{1}}}...\xrightarrow{{{r}_{k}}}v$, the node $v$ should receive the information from node $u$ as well as edges $\left\{ {{r}_{1}},...,{{r}_{k}} \right\}$ in the chain. Here, TrustGNN uses a RotatE-like method~\cite{rotate} to compose the attributes of node $u$ and edges $\left\{ {{r}_{1}},...,{{r}_{k}} \right\}$ in the complex plane\footnote{We truncate a vector ($h$ or $r$) into two as the real and imaginary parts 
respectively.}
\begin{equation}\label{eq:rotate}
    h_{v}^{p}={{h}_{u}}\circ {{r}_{1}}\circ ...\circ {{r}_{k}}
\end{equation}
where ${{h}_{u}},{{r}_{1}},...{{r}_{k}}\in {{\mathbb{C}}^{d}}$, ${{\mathbb{C}}^{d}}$ denotes the complex plane, the modulus $\left| {{r}_{i}} \right|=1$ and $\circ $ is Hadamard (element-wise) product. Due to the propagation property of RotatE, the Hadamard product among edges (${{r}_{1}}\circ ...\circ {{r}_{k}}$) can be seen as the new type of composed relationship specific to the trust chain $p$, and $h_{v}^{p}$ is the information that node $v$ receives on the chain $p$.

Trust is asymmetric. It is worth noting that node $v$ also serves as a trustor node in a trust graph, and a comprehensive node embedding should contain information from its trustee role and trustor role. Thus, for the same trust chain $p$ with node $v$ being the head ($p:v\xrightarrow{{{r}_{1}}}...\xrightarrow{{{r}_{k}}}u$), TrustGNN computes
\begin{equation}\label{eq:rotate-reverse}
    \bar{h}_{v}^{p}={{h}_{u}}\circ {{\bar{r}}_{k}}\circ ...\circ {{\bar{r}}_{1}}
\end{equation}
where ${{\bar{r}}_{i}}$ is conjugate with ${{r}_{i}}$ in Formula~\ref{eq:rotate}. Due to the inversion property of RotatE, the conjugate vector ${{\bar{r}}_{i}}$ can present the inverse relationship of ${{r}_{i}}$ (trust and trusted), and $\bar{h}_{v}^{p}$ is the information that node $v$ receives when node $v$ serves as a trustor node in trust chain $p$.

\textbf{Discussions:} 1) For the nodes in a trust chain, TrustGNN only considers the attributes of the head/tail node but ignores the nodes in between. This is because we just want to capture the trust relationship between head and tail nodes through a trust chain. Other nodes in the chain may be the head/tail node in other chains and thus can be computed to create trust relationships in other propagation processes. 2) RotatE~\cite{rotate} is a knowledge graph (KG) embedding method. It maps each entity and relation in a KG to the complex vector space and defines each relation as a rotation of source entity to target entity. The propagation property of RotatE means that two relations can be composed (Hadamard product) to generate a new type of relations, which can be thought of as a new rotation in complex plane. TrustGNN extends RotatE to compose multiple relations (edges) to generate more complex relations. The inversion property of RotatE means that two relations with opposite semantics can be represented by conjugate vectors in the complex plane. TrustGNN uses the inversion property to compute propagative information from views of both trustor and trustee of a node. Moreover, when TrustGNN computes Formula~\ref{eq:rotate} and Formula~\ref{eq:rotate-reverse} from left to right, the intermediate results exactly correspond to the representation of intermediate nodes of a chain. This is another reason why we can ignore the nodes inner a chain.

\subsection{Trust Chain Based Aggregation}

There are often multiple different types of trust chains in a trust graph and these chains interact to provide evidence for us to create a new trust relationship between the two nodes. Accordingly, a node aggregates information from several chains with different types in our TrustGNN framework. TrustGNN uses a learnable attention score~\cite{gat, attention2} to distinguish the contribution of different types of chains. Before computing the attention score, TrustGNN first summarizes the information on the chains sharing the same type. Consider a node $v$ and a set chains in the $j$-th type, TrustGNN aggregates the information via a sum operation
\begin{equation}
    h_{v}^{{{P}_{j}}}={{W}_{{{P}_{j}}}}\cdot \left({{h}_{v}}+\sum\limits_{\psi \left( p \right)={{P}_{j}}}{h_{v}^{p}}\right)
\end{equation}
where $\psi (\cdot )$ is a mapping function to indicate which type a trust chain is, ${{P}_{j}}$ the $j$-th type of trust chain, and ${{W}_{{{P}_{j}}}}$ a learnable parameter matrix specific to ${{P}_{j}}$. $h_{v}^{{{P}_{j}}}$ is the representation of node $v$ for the $j$-th chain type. Correspondingly, we use ${{H}^{{{P}_{j}}}}$ to represent all nodes. Here, a sum operation means TrustGNN treats chains with the same type equally, because chains with the same type have the same propagative process (Formula~\ref{eq:rotate}) and thus should have equal contribution for node embeddings.

Then, for the summarized information ${{H}^{{{P}_{1}}}},{{H}^{{{P}_{2}}}},...,{{H}^{{{P}_{j}}}}$, TrustGNN aggregates them with different weights as chains with different types have different impact on creating trust and, moreover, there is usually a complex interaction between different types of chains. TrustGNN learns weights through an discriminative attention mechanism~\cite{gat, attention2} to model the composable nature in trust graphs. Specifically, TrustGNN first transforms chain-type-specific representation (${{h}^{{{P}_{j}}}}$) through a nonlinear transformation, and then measures the score of the chain-type-specific representation as the similarity of transformed representation with a chain-type-level attention vector $q$. Furthermore, TrustGNN averages the score of all the chain-type-specific node representations, which can be explained as the importance of each chain type
\begin{equation}
    {{w}_{{{P}_{j}}}}=\frac{1}{\left| \mathcal{V} \right|}\sum\limits_{v\in \mathcal{V}}{{{q}^{T}}\cdot \tanh \left( {{W}_{attn}}\cdot h_{v}^{{{P}_{j}}}+b \right)}
\end{equation}
where ${{W}_{attn}}$ is a learnable parameter matrix, $b$ a bias vector, $q$ the chain-type-level vector of attention. The parameter ${{W}_{attn}}$ and $b$ are shared for all chain types. Then, the softmax function is adopted to normalize the score
\begin{equation}\label{eq:attn-norm}
    {{\alpha }_{{{P}_{j}}}}=\frac{\exp ({{w}_{{{P}_{j}}}})}{\sum\nolimits_{c=1}^{k}{\exp ({{w}_{{{P}_{c}}}})}}
\end{equation}
which can be interpreted as the contribution of the chain type ${{P}_{j}}$. The higher the weight of ${{P}_{j}}$, the more contribution ${{P}_{j}}$ provides in the composable nature. TrustGNN aggregates these chain-type-specific representations to obtain the embeddings
\begin{equation}
    Z=\sum\nolimits_{j=1}^{k}{{{\alpha }_{{{P}_{j}}}}\cdot {{H}^{{{P}_{j}}}}}
\end{equation}

Note that in Section~\ref{sec:method-prop}, TrustGNN computes node representations from two aspects: the trustee role and the trustor role. The embedding $Z$ only aggregates the information about the trustee role. Therefore, TrustGNN needs to compute an embedding $\bar{Z}$ for the trustor role and combines the two to get a more comprehensive node embedding
\begin{equation}
    \bar{Z}=\sum\nolimits_{j=1}^{k}{{{{\bar{\alpha }}}_{{{P}_{j}}}}\cdot {{{\bar{H}}}^{{{P}_{j}}}}}
\end{equation}
where ${{\bar{H}}^{{{P}_{j}}}}$ is the node representations about the trustee role on the ${{P}_{j}}$ chain type, and ${{\bar{\alpha }}_{{{P}_{j}}}}$ is the corresponding weight computed by Formula~\ref{eq:attn-norm}. Note that ${{\bar{\alpha }}_{{{P}_{j}}}}$ and ${{\alpha }_{{{P}_{j}}}}$ are irrelevant because TrustGNN uses another parameters to compute ${{\bar{\alpha }}_{{{P}_{j}}}}$, i.e., ${{\bar{W}}_{attn}}$ and $\bar{b}$. This means that a chain will have different contributions when it serves for embeddings about different roles (trustee or trustor). TrustGNN finally combines $Z$ and $\bar{Z}$ to obtain the final embedding
\begin{equation}
    {{Z}_{final}}=W\cdot \left( Z\left\| {\bar{Z}} \right. \right)
\end{equation}
where $||$ is the concatenation operation and $W$ a learnable parameter matrix to transform the embeddings into a low-dimensional space.

\subsection{The Predictor Layer}
TrustGNN is a trust evaluation model based on node embeddings. Specifically, given two nodes $u$ and $v$ as a trustor-trustee pair, TrustGNN predicts their trust relationships based on their embeddings
\begin{equation}
    {{\tilde{y}}_{u-v}}=\sigma \left(\text{MLP}({{z}_{u}}||{{z}_{v}})\right)
\end{equation}
where ${{z}_{u}}$ and ${{z}_{v}}$ are node embeddings (two vectors in ${{Z}_{final}}$), $||$ is the concatenation operation, $\text{MLP}(\cdot )$ is a multi-layer perceptron (MLP), and $\sigma(\cdot)$ is an activation function. Note that because of asymmetric property of trust relationships in trust graphs, ${{\tilde{y}}_{u-v}}\ne {{\tilde{y}}_{v-u}}$. As the number of trust types in trust graphs is limited, the trust evaluation task is equivalent to the classification task based on embedding pairs. TrustGNN is a semi-supervised model and it minimizes cross-entropy between the predicted values and ground-truth values as the loss function
\begin{equation}
    \mathcal{L}=\text{cross}\_\text{entropy}(Y,\tilde{Y})
\end{equation}
where $Y$ is the observed categorical values in the trust graph and $\tilde{Y}$ the predicted result by TrustGNN. TrustGNN is an end-to-end model, as the parameters in predictor layer and that in embedding module are updated together via back propagation, under the guidance of this unified objective function.

\section{Experiments}\label{sec:Experiments}
In this section, we first provide the experimental setup, including dataset description, baselines for comparison, and parameters and model settings. Then we show the comparison results with baseline methods and finally give the ablation experiments and parameter analysis.

\subsection{Experiment Setup}

\subsubsection{Datasets} We adopt two widely-used, real-world datasets for model evaluation (as done in most related works such as Guardian~\cite{Guardian}, NeuralWalk~\cite{NeuralWalk} and OpinionWalk~\cite{OpinionWalk}). The statistics of the two datasets are shown in Table~\ref{tab:datasets}.
\begin{enumerate}
    \item Advogato is a dataset collected from the online software development community, in which a trust edge from user P to user Q represents P's trust in Q, representing the ability of user Q in software development. The trust relationships can be divided into four types, i.e., \{Observer, Apprentice, Journeyer, Master\}. Nodes and edges in Advogato have no attributes so that we initialize node attributes as random vectors and treat these vectors as a part of learnable parameters of the neural networks.
    \item Pretty-Good-Privacy (PGP) is a dataset obtained from the public key certification network, in which the trust edge from user P to Q represents that user P attests to user Q's trust. Similarly, trust in the this dataset contains four different levels of trust and attributes are initialized by neural networks.
\end{enumerate}

\begin{table}[t]
\caption{Statistics of datasets.}\label{tab:datasets}
\centering
\resizebox{1.0\linewidth}{!}{
\begin{tabular}{c|c|c|c|c}
\toprule[1.5pt]
Datasets  & Nodes & Edges  & Diameter &Avg. Degree  \\ \midrule[0.6pt]
Advogato & 6,541  & 51,127  &4.82         & 19.2     \\
PGP      & 38,546 & 317,979 & 7.7        & 16.5      \\ \bottomrule[1.5pt]
\end{tabular}
}
\end{table}

\subsubsection{Baselines} We compare the proposed TrustGNN with four state-of-the-art methods, corresponding to four popular types of trust evaluation methods, i.e., a matrix factorization-based method, a walk-based method, a deep neural network-based method, and a GNN-based method.
\begin{enumerate}
    \item Matri~\cite{MATRI} uses matrix factorization to infer trust relationships between users. It utilizes locally generated trust relationships to characterize multiple complex latent factors between each trustor and trustee, and further introduces prior knowledge and trust propagation to improve prediction accuracy. The trust relationship between each node pair is captured by computing the similarity between the trustor's latent vector and the trustee's latent vector in a learnable latent space.
    \item OpinionWalk~\cite{OpinionWalk} uses random walk to model the trust relationship between users. The method uses the breadth-first search method to create the trust relationship for every two users who do not have a direct trust relationship, uses Dirichlet distribution to model the data distribution, and expresses the direct trust value between users in the form of a matrix.

    \item NeuralWalk~\cite{NeuralWalk} is a deep neural network based method. It first models the propagation and fusion of direct trust among users in the trust social network by designing a neural network module named WalkNet. Then, with WalkNet as the neural subunit, the multi-hop social trust among users in the network is captured in the form of iterative neural subunits. This approach is the state-of-the-art modelling scheme for trust evaluation, but it typically requires greater computational and memory complexity than other approaches.
    
    \item Guardian~\cite{Guardian} applies GNN to the trust evaluation task for the first time. It designs a trust convolutional layer to model trust interactions in social networks. The method is also designed to incorporate the two trust relationships of popularity and engagement into the model learning process, respectively. It finally predicts the trust relationship of each trustor-trustee pair based on the representations. Note that our proposed TrustGNN is also a graph neural network-based method, while TrustGNN better models the propagative nature and composable nature of trust among users to achieve better performance.
\end{enumerate}

\subsubsection{Evaluation metrics} Follow the metrics used in Guardian~\cite{Guardian}, we adopt two popular metrics to evaluate the effectiveness of the new proposed method, including F1-score and Mean Absolute Error (MAE). In the experiments, we run each method 20 times and take the average of these results as the final result. Note that larger F1-score values indicate better prediction performance, whereas smaller MAE values indicate better prediction performance. For the split of the datasets, since OpinionWalk is deductive, we conduct experiments with randomly selected 1,000 trustor-trustee pairs for both datasets. As for other baselines as well as our TrustGNN, we randomize the two datasets into 80\% trustor-trustee pairs to form the training set and the rest for the test set. When computing F1-score, we map the predicted relationship into categorical values, i.e., \{Observer: 0, Apprentice:1, Journeyer: 2, Master: 3\}. When computing MAE, we follow the Matri~\cite{MATRI} and Opinion~\cite{OpinionWalk} to map the relationship into scalar values, i.e., \{Observer: 0.1, Apprentice:0.4, Journeyer: 0.7, Master: 0.9\}.

\subsubsection{Parameter settings} 
For the baselines, we adopt their default parameter settings as they often lead to the best results. For our TrustGNN, we set learning rate to $0.005$, the dimensions of nodes and edge attributes to $1024$, and the dimension of attention vector ($q$) and representation vector ($h$, $r$, $z$) to $128$. We set the maximum length of trust chains $K$ to $2$ since it is typically sufficient to model trust relationships in online social networks. And when $K$ is too large, there will be additional computational cost and more noise may be introduced. We implemented our proposed method using Python-3.7\footnote{https://www.python.org/} and Pytorch-1.6\footnote{https://pytorch.org/}. All the experiments were conducted on the same machine with Linux system (Ubuntu 5.4.0), Intel(R) Xeon(R) CPU E5-2680, 128GB RAM and 2 NVIDIA 1080Ti GPUs.

\begin{table}[t]
\caption{Performance Comparision with Baselines. Larger values of F1-score (and smaller values of MAE) indicate better results. The best results are in bold.}\label{tab:comp}
\centering
\resizebox{1.0\linewidth}{!}{
\begin{tabular}{c|cc|cc}
\toprule[1.5pt]
\multirow{2}{*}{Methods} & \multicolumn{2}{c|}{Advogato}    & \multicolumn{2}{c}{PGP}         \\ \cmidrule(r){2-3} \cmidrule(r){4-5}
                        & F1-Score        & MAE            & F1-Score        & MAE            \\ \midrule[0.6pt]
TrustGNN                & \textbf{74.4\%} & \textbf{0.081} & \textbf{87.2\%} & \textbf{0.083} \\ \midrule[0.6pt]
Guardian                & 73.0\%          & 0.087          & 86.7\%          & 0.086          \\ \midrule[0.6pt]
NeuralWalk              & 74.0\%          & 0.082          & --              & --             \\ \midrule[0.6pt]
OpinionWalk             & 63.3\%          & 0.232          & 66.8\%          & 0.251          \\ \midrule[0.6pt]
Matri                   & 65.0\%          & 0.141          & 67.3\%          & 0.136          \\ \bottomrule[1.5pt]
\end{tabular}
}
\end{table}
\subsection{Performance Comparisons}
We first evaluate the performance of TrustGNN on Advogato dataset. The experimental results are provided in Table~\ref{tab:comp} where the best results are in bold. The proposed TrustGNN outperformed all baselines. Compare to the second best method, i.e., NeuralWalk, TrustGNN improves 0.4\% in terms of F1-Score. Correspondingly, the error rate reduces by 1.9\%. In terms of MAE, TrustGNN also achieves the best performance, equivalent to the best baseline NeuralWalk. We also evaluate TrustGNN on the larger PGP dataset and the results are provided in Table~\ref{tab:comp}. Compared to Guardian, TrustGNN improves 0.5\% and reduces errors by 3.8\% in terms of micor-F1, and also achieves the best MAE score. Also of note is that, because of the huge computational and memory complexity of NeuralWalk, it fails to get the results on PGP dataset (as the number of nodes/edges in PGP is $\sim 6$ times in Advogato). In contrast, TrustGNN is lightweight and can meanwhile achieve the best performance.

Besides, as shown in Table~\ref{tab:comp}, there is a great gap between neural network-based methods (TrustGNN, Guardian, NeuralWalk) and the others (Opinion, Matri). This means that the powerful learning ability of neural networks allows it to better solve the trust evaluation task. In the above three neural networks-based methods, the newly proposed TrustGNN has more elaborate designs. TrustGNN follows the principle of knowledge graph embedding to propagate information on chains, which can better model the propagative nature of social trust. It also applies attention mechanism between different types of chains, which is better to capture the composable nature to distinguish the contributions of different type of chains when creating new trust relationships.

We also evaluate TrustGNN on these two datasets with different training and testing splits. Specifically, we set the portions of training set as 40\%, 60\%, 80\% and report the performance in terms of F1-Score and MAE. We select Guardian and Matri as comparision algorithms. The results are shown in Table~\ref{tab:split-advogato} and Table~\ref{tab:split-pgp}. Obviously, TrustGNN achieves best result across all splits, on both Advogato and PGP datasets. This further demonstrated the robustness and stability of TrustGNN model. 

\begin{table}[t]
\caption{Performance evaluation with different training sizes on Advogato. The best results are in bold.}\label{tab:split-advogato}
\centering
\resizebox{1.0\linewidth}{!}{
\begin{tabular}{c|ccc}
\toprule[1.5pt]
Methods                    & \begin{tabular}[c]{@{}c@{}}Training Set\\ (\%)\end{tabular} & F1-Score          & MAE              \\ \midrule[0.6pt]
\multirow{3}{*}{TrustGNN} & 80\%                                                        & \textbf{74.4\%$\pm $0.1\%} & \textbf{0.081$\pm $0.001} \\ 
                          & 60\%                                                        & \textbf{72.6\%$\pm $0.1\%} & \textbf{0.088$\pm $0.001} \\ 
                          & 40\%                                                        & \textbf{70.1\%$\pm $0.1\%} & \textbf{0.096$\pm $0.001} \\ \midrule[0.6pt]
\multirow{3}{*}{Guardian} & 80\%                                                        & 73.0\%$\pm $0.1\% & 0.087$\pm $0.001 \\ 
                          & 60\%                                                        & 71.7\%$\pm $0.2\% & 0.091$\pm $0.001 \\ 
                          & 40\%                                                        & 69.7\%$\pm $0.0\% & 0.098$\pm $0.000 \\ \midrule[0.6pt]
\multirow{3}{*}{Matri}    & 80\%                                                        & 65.0\%$\pm $0.4\% & 0.141$\pm $0.001 \\ 
                          & 60\%                                                        & 63.9\%$\pm $0.3\% & 0.145$\pm $0.001 \\  
                          & 40\%                                                        & 61.7\%$\pm $0.3\% & 0.153$\pm $0.001 \\ \bottomrule[1.5pt]
\end{tabular}
}
\end{table}

\begin{table}[h]
\caption{Performance evaluation with different training sizes on PGP. The best results are in bold.}\label{tab:split-pgp}
\centering
\resizebox{1.0\linewidth}{!}{
\begin{tabular}{c|ccc}
\toprule[1.5pt]
Methods                             & \begin{tabular}[c]{@{}c@{}}Training Set\\ (\%)\end{tabular} & F1-Score          & MAE               \\ \midrule[0.6pt]
\multirow{3}{*}{TrustGNN} & 80\%                                                        & \textbf{87.2\%$\pm $0.1\%} & \textbf{0.083$\pm $0.001}  \\  
                                   & 60\%                                                        & \textbf{86.3\%$\pm $0.1\%} & \textbf{0.090$\pm $0.001}  \\  
                                   & 40\%                                                        & \textbf{85.4\%$\pm $0.1\%} & \textbf{0.097$\pm $0.001}  \\ \midrule[0.6pt]
\multirow{3}{*}{Guardian} & 80\%                                                        & 86.7\%$\pm $0.1\% & 0.086$\pm $0.001  \\ 
                                   & 60\%                                                        & 85.9\%$\pm $0.1\% & 0.091$\pm $0.001  \\ 
                                   & 40\%                                                        & 84.6\%$\pm $0.1\% & 0.100$\pm $0.001  \\ \midrule[0.6pt]
\multirow{3}{*}{Matri}    & 80\%                                                        & 67.3\%$\pm $0.7\% & 0.136$\pm $0.0003 \\ 
                                   & 60\%                                                        & 64.7\%$\pm $0.1\% & 0.141$\pm $0.0004 \\  
                                   & 40\%                                                        & 60.5\%$\pm $0.1\% & 0.164$\pm $0.0001 \\ \bottomrule[1.5pt]
\end{tabular}
}
\end{table}

\subsection{Visualization and Interpretation} 
\begin{figure}[t]  
  \centering       
  \includegraphics[width=0.87\linewidth]{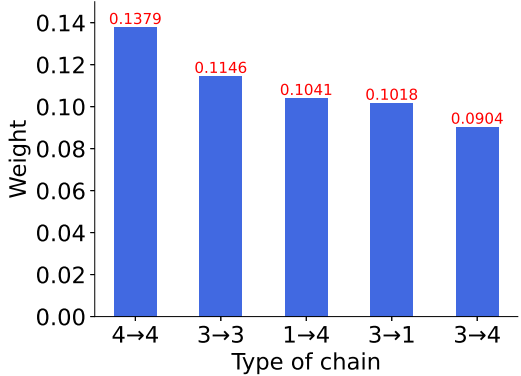} 
  \caption{Visualization of top 5 types of chains. We limit the length of chains to 2. For example, “1→4” means the chain type that a trustor trusts an intermediate node with a trust value of 1 and the intermediate node trusts a trustee with a trust value of 4.}\label{fig:chain-weight}  
\end{figure}

We visualize the weight of each type of trust chains in TrustGNN, i.e., $\alpha$ in Formula~\ref{eq:attn-norm}. TrustGNN uses attention mechanism to compute contribution of each type of chains. In the experiments, we limit the length of chains to $2$. Both Advogato and PGP datasets have $4$ different types of relationships. Therefore, the number of chain types is ${{4}^{2}}=16$. Here we use Advogato dataset as an example and visualize the top 5 chain types in Fig.~\ref{fig:chain-weight}. We use scalars to represent different types of relationships \{Observer: 1, Apprentice:2, Journeyer: 3, Master: 4\} so that large values indicate strong trust relationships. From Fig.~\ref{fig:chain-weight} we can see that the top 5 types of chains contain relationships with large values. This is reasonable because strong trust relationships are often more helpful to creating new trust relationships. Moreover, TrustGNN accurately assigns the highest weight for the chain type with two strongest relationships ($4\to 4$). This demonstrates that TrustGNN is consistent with the laws of reality and has good interpretation. Note that there are also some relationships with small values in the top 5 types of chains. This may be because trustworthiness is imbalanced in trust graphs. The number of relationships with values of $1$ and $2$ is much larger than relationships with values of $3$ and $4$. The model may be biased due to imbalanced data, but in general, TrustGNN can learn high weight for chains with strong trust relationships.

\subsection{Analysis of Attribute Initialization Methods} As some datasets may not have raw attributes, we need to define the attributes of nodes and edges in advance. In Guardian~\cite{Guardian}, they use node2vec~\cite{node2vec} embeddings and one-hot vectors as node/edge attributes. We follow the Guardian's setting to study what types of attributes are more suitable for our TrustGNN. In this experiment, node attributes are defined as embeddings pre-trained from node2vec or initialized as a set of learnable parameters of the neural networks. Edge attributes are defined as one-hot vectors or initialized as a set of learnable parameters of the neural networks. We test all the cases and report the F1-Score and MAE score in Table~\ref{tab:attr-ablation}. From Table~\ref{tab:attr-ablation} we can find that TrustGNN achieves the best results when node and edge attributes are both initialized as learnable parameters. This may be because the learning ability of neural networks can learn good attributes for nodes and edges. We also noticed that using node2vec embeddings will reduce the performance while using one-hot vectors has little impact on the performance. The one-hot vectors assign discriminative attributes for different types of edges so that it can still distinguish the role of different types of trustworthiness. On the other hand, because node2vec is proposed for graph with homogeneous edges, it may be not so suitable for trust graphs with multi-relations.

\begin{table}[th]
\caption{The results of TrustGNN with different types of attributes.}\label{tab:attr-ablation}
\centering
\resizebox{1.0\linewidth}{!}{
\begin{tabular}{cc|cc|cc}
\toprule[1.5pt]
\multirow{2}{*}{\begin{tabular}[c]{@{}c@{}}Node\\ Attributes\end{tabular}} & \multirow{2}{*}{\begin{tabular}[c]{@{}c@{}}Edge\\ Attributes\end{tabular}} & \multicolumn{2}{c|}{Advogato} & \multicolumn{2}{c}{PGP}              \\ \cmidrule(r){3-4} \cmidrule(r){5-6}
                                                                          &                                                                           & F1-Score        & MAE         & \multicolumn{1}{c}{F1-Score} & MAE   \\ \midrule[0.6pt]
Parameter                                                                 & Parameter                                                                 & 74.4\%          & 0.081       & 87.2\%                        & 0.083 \\ \midrule[0.6pt]
Parameter                                                                 & One-hot                                                                   & 74.2\%          & 0.083       & 87.1\%                        & 0.084 \\ \midrule[0.6pt]
Node2vec                                                                  & Parameter                                                                 & 72.4\%          & 0.088       & 83.3\%                        & 0.114 \\ \midrule[0.6pt]
Node2vec                                                                  & One-hot                                                                   & 72.4\%          & 0.088       & 82.0\%                        & 0.124 \\ \bottomrule[1.5pt]
\end{tabular}
}
\end{table}

\subsection{Ablation Study}

Similar to most deep learning methods, TrustGNN consists of several different components that may have important impact on the model performance. To provide intuitive understanding to the model’s components, we perform experiments comparing TrustGNN with its three variants. The variants are defined as follows: 1) In the trust chain based propagation process (Section~\ref{sec:method-prop}), a node will receive two kinds of information corresponding to its two roles. Here we ignore the trustor role of nodes and make a node only receive the information from its trustee role, named as TrustGNN-1; 2) We also ignore the trustee role of nodes and make a node only receive the information from its trustor role, named as TrustGNN-2; 3) In trust chain based aggregation, we employ a sum operation instead of discriminative aggregation, named as TrustGNN-3. We report the average F1-Score at different training ratios for comparison.

\begin{figure}[th]
  \centering       
    \subfloat[Advogato] 
    {
        \begin{minipage}[t]{0.25\textwidth}
            \centering          
            \includegraphics[width=1.0\textwidth]{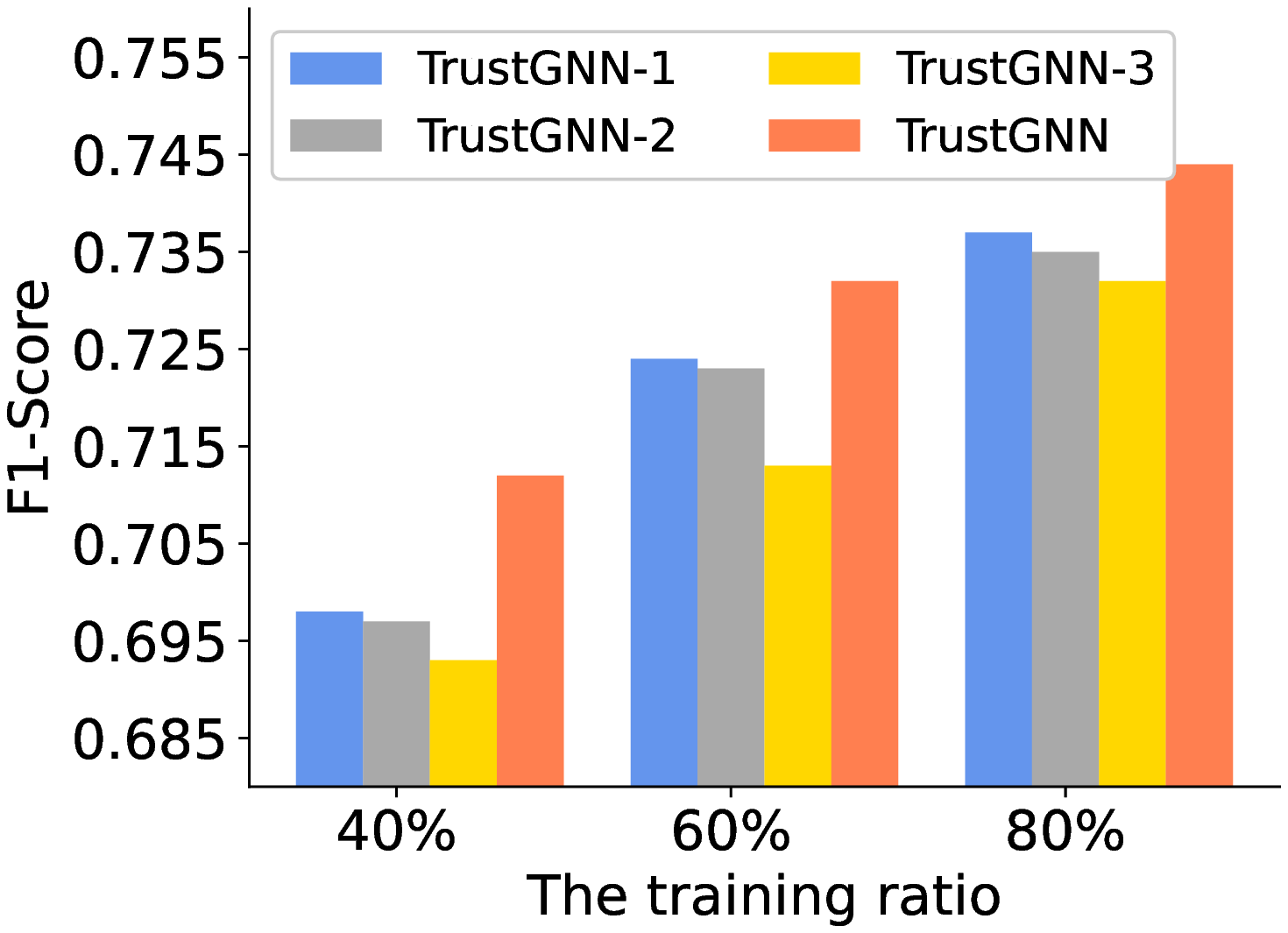}  
        \end{minipage}%
    }
    \subfloat[PGP] 
    {
        \begin{minipage}[t]{0.25\textwidth}
            \centering      
            \includegraphics[width=1.0\textwidth]{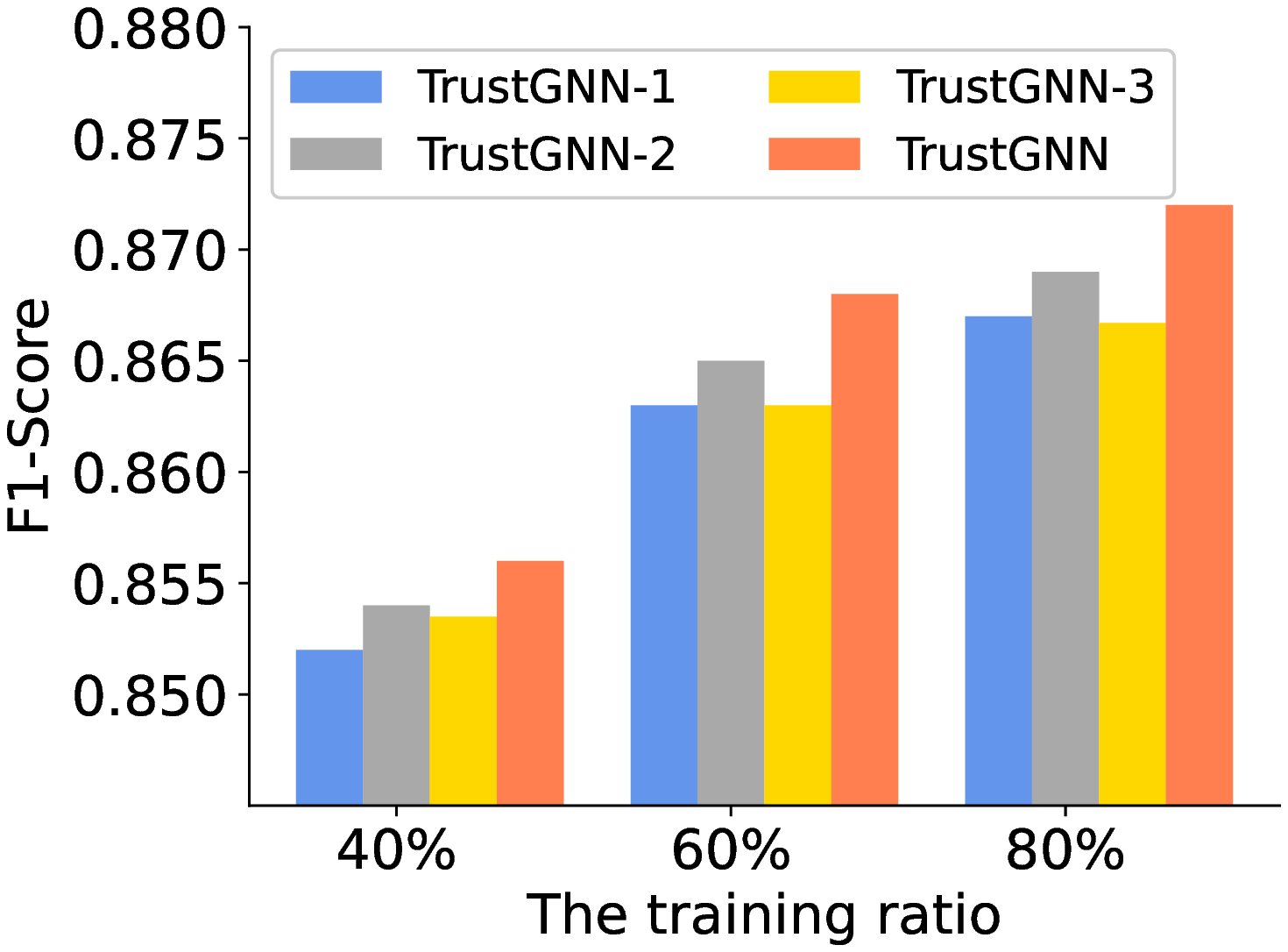}   
        \end{minipage}
    }%
  \caption{Comparisons of F1-Score of our TrustGNN and three variants at different training ratios on two datasets.}\label{fig:xr}   
\end{figure}

As shown in Fig.~\ref{fig:xr}, we can draw three conclusions: (1) TrustGNN consistently outperforms all its variants on both datasets, illustrating the effectiveness of simultaneously considering two different roles of target nodes and discriminative aggregation of different trust chains. (2) TrustGNN-1 and TrustGNN-2 have different performance advantages in different datasets, which further illustrates the necessity of comprehensively considering two different roles of target nodes. (3) TrustGNN-3 is consistently lower than the other three baselines, which illustrates the rationality of aggregation considering the importance of different trust chains, and can more fully capture the composable nature of trust.

\begin{figure}[t]  
  \centering       
    \subfloat[Advogato] 
    {
        \begin{minipage}[t]{0.25\textwidth}
            \centering          
            \includegraphics[width=1.0\textwidth]{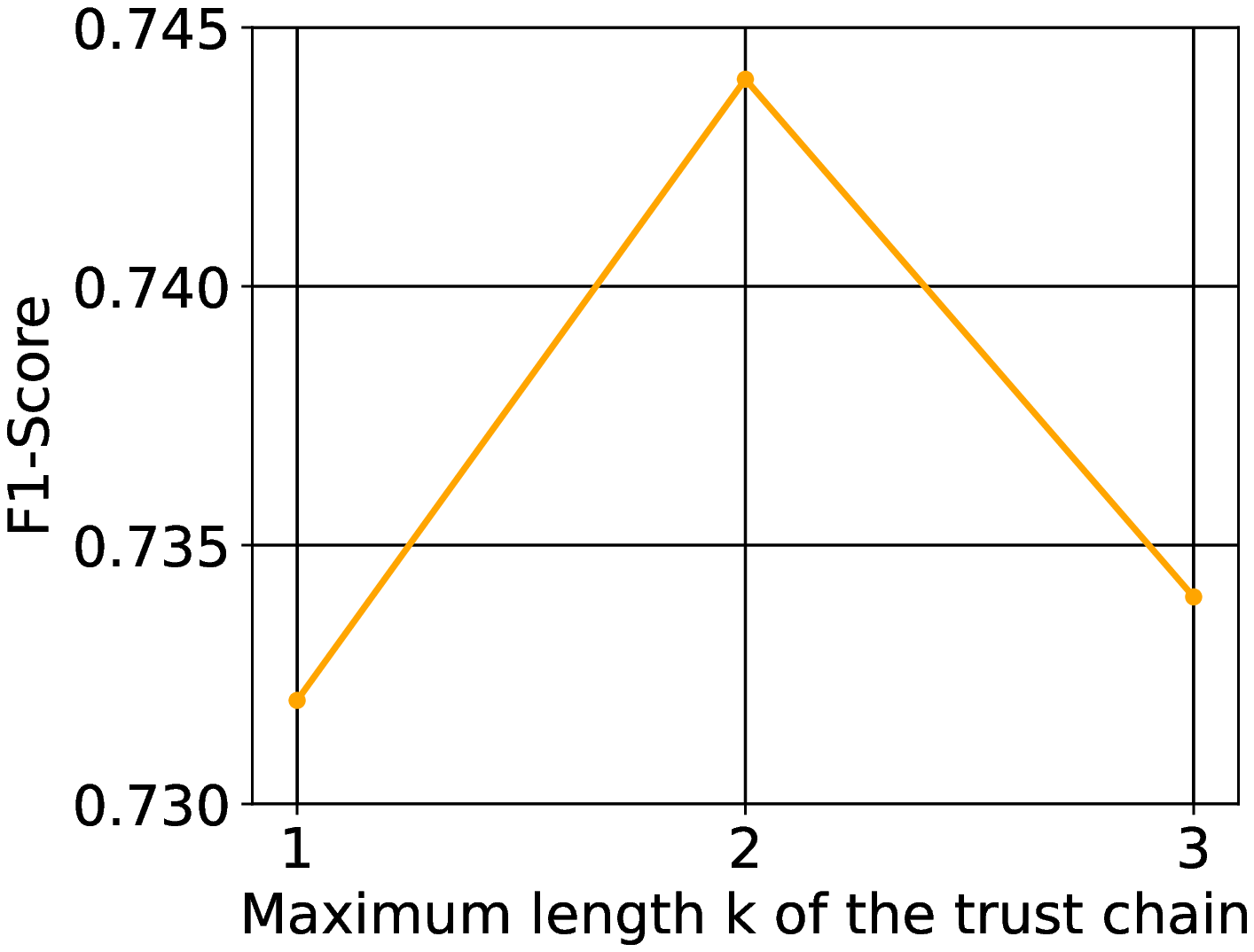}  
        \end{minipage}%
    }
    \subfloat[PGP] 
    {
        \begin{minipage}[t]{0.25\textwidth}
            \centering      
            \includegraphics[width=1.0\textwidth]{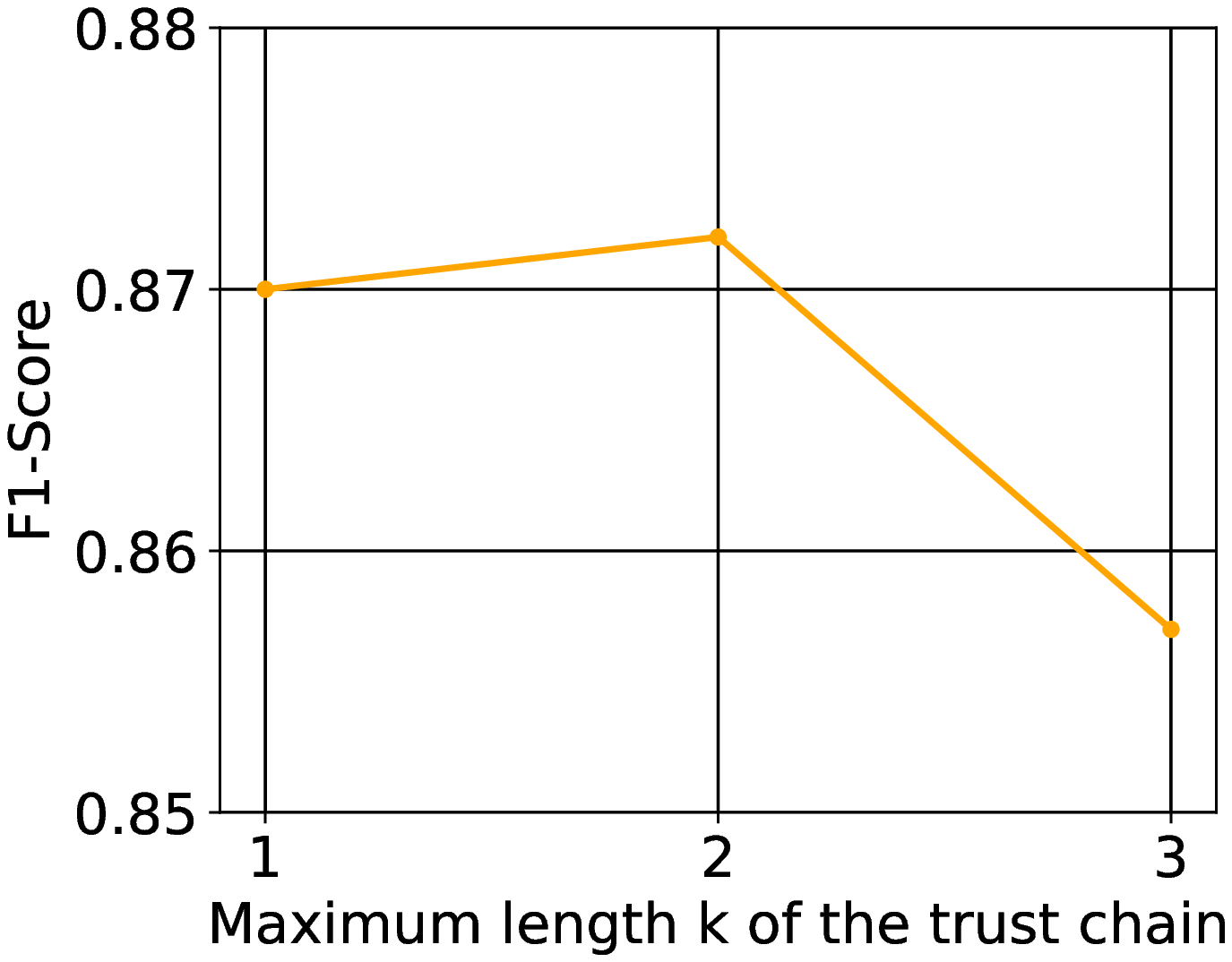}   
        \end{minipage}
    }%
  \caption{Parameter analysis of maximum length of trust chain $K$ on two datasets.}\label{fig:4}  
\end{figure}

\subsection{Parameter Analysis}
We further analyzed how model parameters affected TrustGNN performance. The most important hyperparameters of TrustGNN are the maximum length of trust chain $K$ and the attribute dimension of edge/node. Here we report the average F1-Score on Advogato and PGP datasets.

\textbf{Maximum length of trust chain.} In order to analyze the influence of the maximum length of trust chain $K$ on the model performance, we test the performance when $K=1, 2,$ and $3$ on two datasets respectively. As shown in Fig.~\ref{fig:4}, the model performance first increases and then decreases with increase of $K$ on both datasets. This observation is reasonable because trust chains that are too short cannot adequately capture the propagative nature of trust, while trust chains that are too long may introduce more noise as trust will decrease in the propagation process.

\begin{figure}[t]  
  \centering       
    \subfloat[Advogato] 
    {
        \begin{minipage}[t]{0.25\textwidth}
            \centering          
            \includegraphics[width=1.0\textwidth]{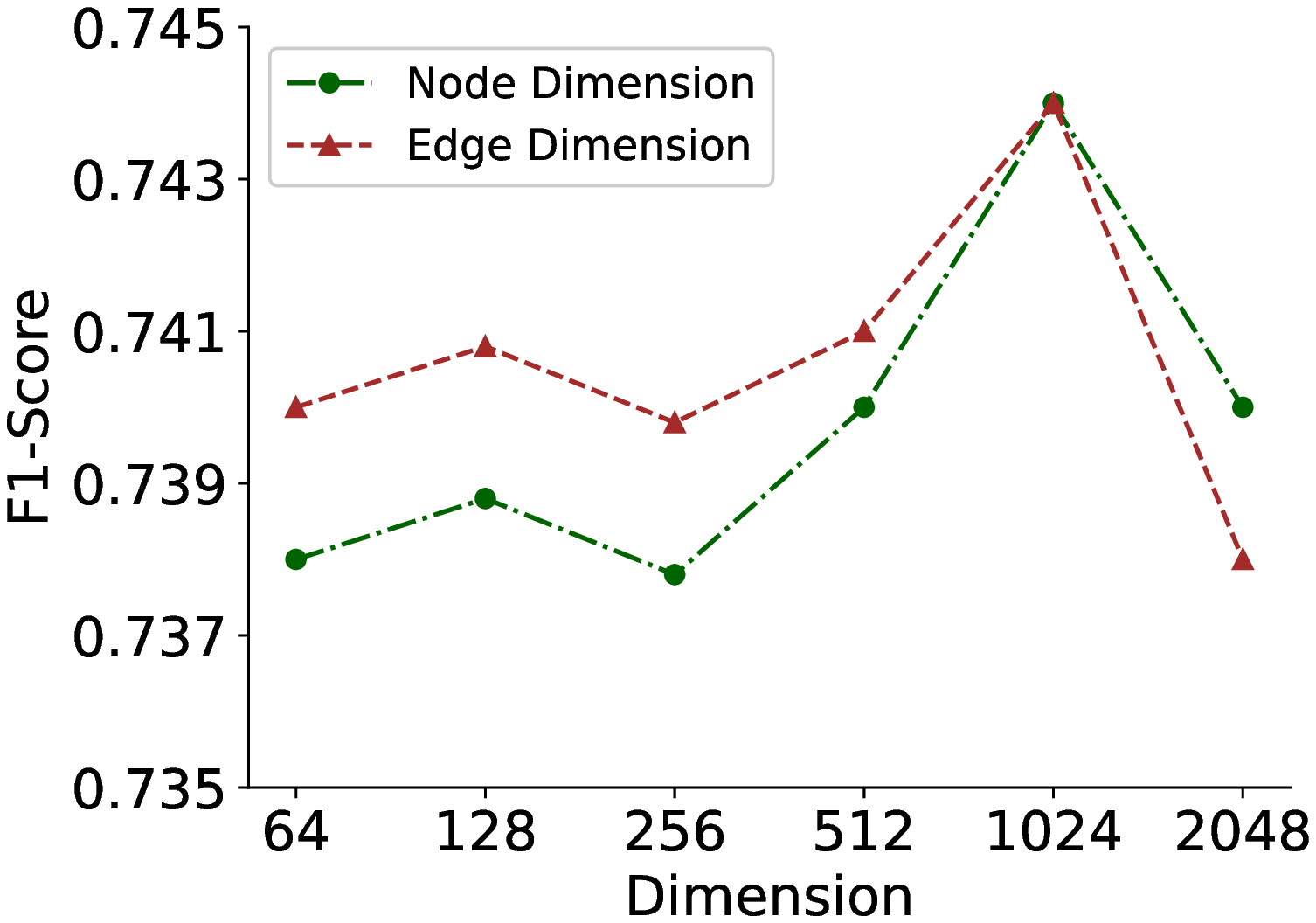}  
        \end{minipage}%
    }
    \subfloat[PGP] 
    {
        \begin{minipage}[t]{0.25\textwidth}
            \centering      
            \includegraphics[width=1.0\textwidth]{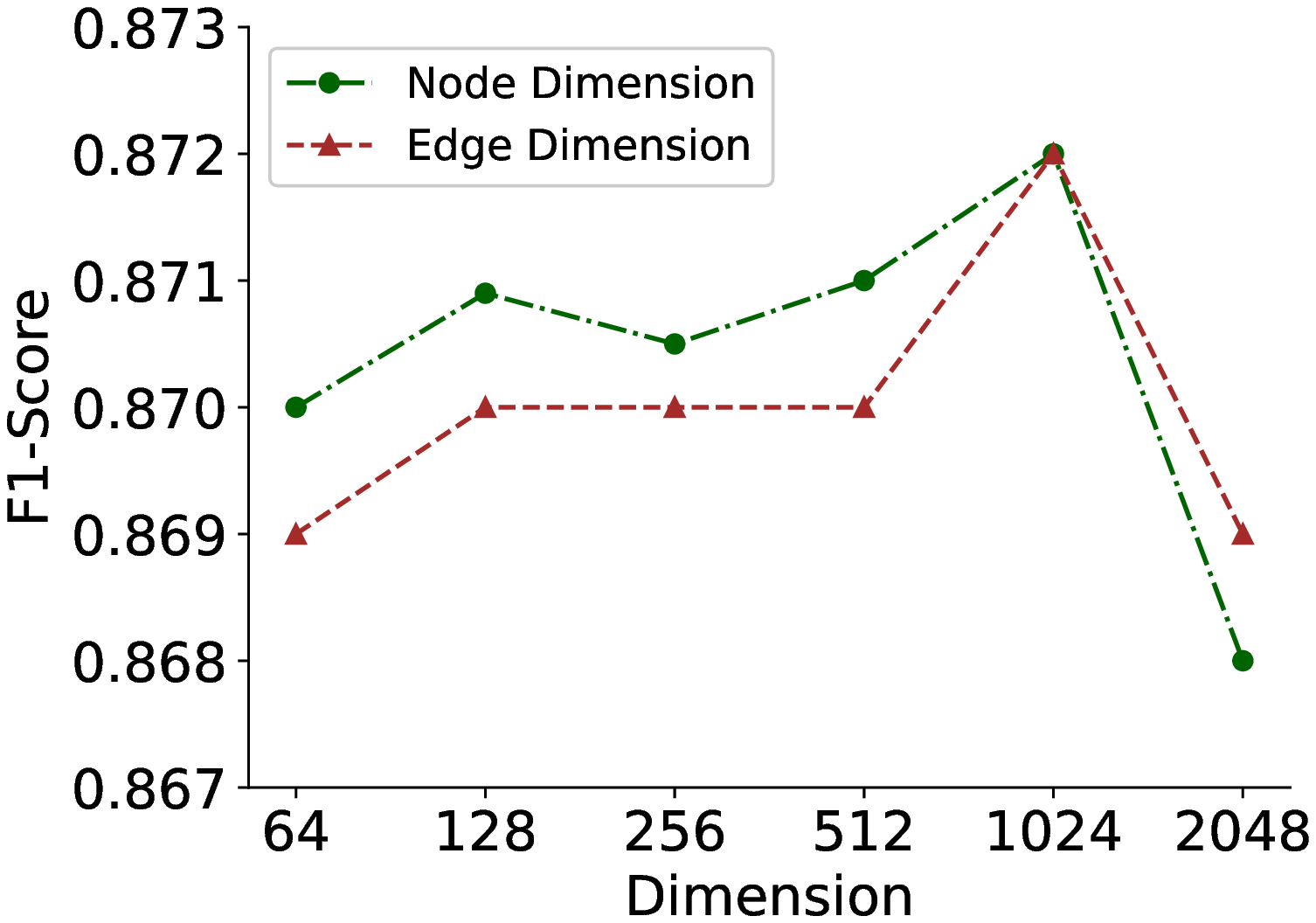}   
        \end{minipage}
    }%
  \caption{Parameter analysis of attribute dimensions of nodes and edges. The default dimension is 1024 for both nodes and edges. We fix the default setting for nodes/edges and analyze another.}\label{fig:analyse-dimension}  
\end{figure}

\textbf{Attribute dimension.} We also analyze the impact of attribute dimension when TrustGNN initializes node and edge attributes as learnable parameters. We fix the dimension of node/edge to $1024$ and test the performance with edge/node dimension varying in $\{64, 128, 256, 512, 1024, 2048\}$. The analysis result of node attribute dimension and that of edge attribute dimension are both shown in Fig.~\ref{fig:analyse-dimension}. They show the same trend. The curves first rise and then fall and TrustGNN achieves best performance when dimensions of nodes and edges are both around $1024$. This may because when the dimension is too small, e.g., $64$, the expressive ability of the model will be weak due to the overly small number of parameters. When the dimension is too large, it is difficult for the model to converge to a good state due to the overly large number of parameters.

\section{Related Work}\label{sec:relatedWork}
\subsection{Trust Evaluation}
Due to the significance of trust in online social networks, trust evaluation has been widely studied and reviewed~\cite{survey1,survey2}. Trust relationships can be formalized as subjective trust measures. The subjective logic-based methods introduce the uncertainty inference process for the subjective nature of trust. For example, TNA-SL~\cite{TNA-SL} simplifies the complex trust graph to a series-parallel graph by deleting the most uncertain path to obtain a canonical graph. Trust measures are expressed as beliefs, and subjective logic is used to calculate the trust between any parties in the network. 3VSL~\cite{3VSL} assumes social trust among users is determined by the objective evidences, and believes that cognitive features of social trust among users are not what they care about. It distinguishes the posteriori and priori uncertainties existing in trust, and the difference between distorting and original opinions, thus be able to compute multi-hop trust in arbitrary graphs. OpinionWalk~\cite{OpinionWalk} uses an opinion matrix to model the topology of the trust graph. Each entry in the matrix is a direct opinion of two corresponding users. They then devised multiple matrix-like operations by using discounting and combining operations instead of traditional multiplication and summation. In this matrix-like operations, the discounting and combining operations are adopted to model trust propagation and fusion.

Another type of methods which is based on probability statistics often uses mathematical distribution to model trust evaluation. For example, Liu \emph{et al}.~\cite{Liu1} propose a context-aware trust model to predict dynamic trust by modeling the interactions of agents using Hidden Markov Model. They first apply information theory and multiple discriminant analysis to select the most useful features and combine the same to generate a compact and effective feature vector, which is viewed as the observations associated with each interaction. Then they propose a HMM based trust model considering such contextual information to capture dynamic behavior of the target agent. Li \emph{et al}.~\cite{Itrust} quantify interpersonal trust by analyzing the frequency of social interactions between users and their friends on Facebook. They adopt bidirectional interaction relationships in online social networks to deconstruct users' social behaviors and apply principal component analysis to estimate interpersonal trust. Liu \emph{et al}.~\cite{Liu2} propose a framework to accurately model single values as parameters. The model first uses a probabilistic graph model to model trustors' opinions and biases to his rating on the trustee. 

Very recently, many trust evaluation methods based on machine learning technology are also constantly being proposed. For example, Matri~\cite{MATRI} is a multi-aspect trust inference model using matrix factorization. It views the trust evaluation problem as a recommendation task, then borrows the rich methodologies from collaborative filtering. NeuralWalk~\cite{NeuralWalk} first designs a neural subunit named WalkNet to model the propagation and fusion of direct trust in trust social networks. Then, the unknown multi-hop trust relationship between users is calculated by iterating this subunit continuously. Especially, with the success of deep learning, GNN-based model Guardian~\cite{Guardian} has been proposed. Guardian divides the neighbor nodes directly connected to the target node into in-degree neighbor nodes and out-degree neighbor nodes, and uses the convolutional layer for information aggregation respectively. Then it obtains high-order neighbor information by stacking multiple convolutional layers. Medley~\cite{Medley} leverages social interactions that change dynamically over time to perform trust evaluations on dynamic online social networks. The method proposes a functional temporal encoding module to model temporal features and employs an attention mechanism to assign higher importance to recent temporal social relationships. 

\subsection{Graph Nerual Networks}
Graph neural networks (GNNs)~\cite{gnnsurvey1, gnnsurvey2} have been proved to be an effective tool for analyzing graph-structural data. Most GNNs learn node representations by aggregating message from neighboring nodes based on the guidance of graph topology. GNNs are first applied to homogeneous graphs. For example, GCN~\cite{gcn} simplifies spectral graph convolutions~\cite{SpectralGCN} by using a localized first-order approximation. GAT~\cite{gat} considers that different neighbors of the target node may have different importance, and integrates the importance coefficients into the aggregation function. GraphSAGE~\cite{graphsage} aggregates the neighbor information of the target node in a learnable manner, and learns node representations in an inductive manner by using node-level sampling methods. R-GCN~\cite{rgcn} designs multiple convolution operations in units of edge types to model the impact of different types of edges on the target node. GIN~\cite{gin} aims to study the expressive power of GNNs by studying the ability to distinguish any two graphs, and proposes a new framework, which is shown to have the same expressive power as Weisfeiler-Lehman. Beyond homogeneous graphs, there are also some GNNs for heterogeneous graphs. Heterogeneous graphs contain more than one type of nodes or edges, which can better model real-world systems. In heterogeneous GNNs, the most important design is using discriminate operations to distinguish the role of different types of nodes/edges in GNN framework. For example, HAN~\cite{han} and MAGNN~\cite{magnn} use attention mechanism to learn the weights for information with different semantic. HetSANN~\cite{hetsann} and HGT~\cite{hgt} proposes a type-aware attention layer to model the relationships between different types of nodes without directly adopting traditional convolutional layers. In addition, GNNs have also been continuously extended to computer vision~\cite{cv,cv2}, natural language processing~\cite{wu2021graph,chen2019reinforcement,xu2018graph2seq} and other fields~\cite{gnnsurvey1}, and have achieved great success.

Despite the great success of previous work, there are seldom GNN-based work for trust evaluation tasks. A trust graph can be seen as a heterogeneous graph but it has more field-related properties. In this paper we show our insights about trust graph and design a new GNN-based trust evaluation method by integrating trust properties into GNN framework. 

\subsection{Knowledge Graph Embedding}
Knowledge graph embedding (KGE) focus on learning low-dimensional embeddings for entities and their relations in knowledge graph (KG)~\cite{kg-emb1, kg-emb2}. The learned embeddings can be used for KG tasks such as KG alignment~\cite{kg-alignment1, kg-alignment2} and relation predictions~\cite{kg-emb2, kg-link}. KGE methods can be viewed as modeling (head entity, relation, tail entity) triples. Concretely, a scoring function is defined to measure the plausibility of triplets given embeddings and help update the representation of the training data. The scoring functions have many designing criteria, such as translation relation~\cite{transe}, rotational relation~\cite{rotate}, inner product~\cite{kg-inner} and others~\cite{kg-other1, kg-other2}. For example, RotatE~\cite{rotate} is a recent proposed method. It maps the entities and relations in a KG to complex vector space, and defines the relations between entities as rotations from source entities to target entities. The interaction between entities and relations in score functions maintains the semantic relationships between entities and relations, and also can be applied to model trust relationships. In this paper we extend design principles in KGE to model the propagative nature of trust, not only for triplets but also for more complex interaction, e.g., multiple user and relationships in a trust chain.

\section{Conclusion}\label{sec:conclusion}
In this paper, we proposed TrustGNN, a new graph neural network (GNN) based method for trust evaluation in online social networks. In TrustGNN, we for the first time, explicitly integrated the propagative and composable nature into GNN framework to learn comprehensive embeddings for better trust evaluation. We defined trust chains to model the propagative pattern of trust and extended knowledge graph embedding methods to model the interaction among nodes and relationships in each trust chain. We further used attention mechanism to learn the importance coefficients of different types of chains, in order to distinguish the contributions of different propagative process. Experimental results on widely-used real-world datasets have demonstrated the superiority of the new proposed TrustGNN. The ablation studies also showed the effectiveness of key designs in TrustGNN. Our proposed TrustGNN has achieved good performance, but deep learning techniques still have great potential for trust evaluation tasks. In the future, we would like to incorporate the dynamic behavior of social networks into our model. We would also like to explore self-supervised techniques and apply them effectively to trust evaluation in social networks.

\bibliographystyle{IEEEtran}
\bibliography{ref}

\end{document}